\documentclass[sigconf,anonymous=false]{acmart}

\usepackage{array}

\copyrightyear{2021}
\acmYear{2021}
\setcopyright{acmcopyright}\acmConference[ICMI '21]{Proceedings of the 2021 International Conference on Multimodal Interaction}{October 18--22, 2021}{Montréal, QC, Canada}
\acmBooktitle{Proceedings of the 2021 International Conference on Multimodal Interaction (ICMI '21), October 18--22, 2021, Montréal, QC, Canada}
\acmPrice{15.00}
\acmDOI{10.1145/3462244.3479893}
\acmISBN{978-1-4503-8481-0/21/10}

\begin{document}

\title{Long-Term, in-the-Wild Study of Feedback\\ about Speech Intelligibility for K-12 Students\\ Attending Class via a Telepresence Robot}
\renewcommand{\shorttitle}{Speech Intelligibility Feedback}


\author{Matthew Rueben}
\affiliation{%
  \institution{New Mexico State University}
%  \department{Intergroup Human-Robot Interaction Lab}
  \city{Las Cruces}
  \state{New Mexico}
  \country{USA}
}
\email{mrueben@nmsu.edu}

\author{Mohammad Syed}
\affiliation{%
  \institution{San Jose State University}
  \city{San Jose}
  \state{California}
  \country{USA}
}
\email{mohammad.syed@sjsu.edu}

\author{Emily London}
\affiliation{%
  \institution{University of Southern California}
  \city{Los Angeles}
  \state{California}
  \country{USA}
}
\email{elondon@usc.edu}

\author{Mark Camarena}
\email{markcama@usc.edu}
\author{Eunsook Shin}
\email{eunsooks@usc.edu}
\affiliation{%
  \institution{University of Southern California}
  \city{Los Angeles}
  \state{California}
  \country{USA}
}

\author{Yulun Zhang}
\email{yulunzha@usc.edu}
\author{Timothy S. Wang}
\email{wangtimo@usc.edu}
\affiliation{%
  \institution{University of Southern California}
  \city{Los Angeles}
  \state{California}
  \country{USA}
}

\author{Thomas R. Groechel}
\email{groechel@usc.edu}
\author{Rhianna Lee}
\email{rhiannal@usc.edu}
\author{Maja J. Matari\'c}
\email{mataric@usc.edu}
\affiliation{%
  \institution{University of Southern California}
  \city{Los Angeles}
  \state{California}
  \country{USA}
}

\renewcommand{\shortauthors}{Rueben et al.}

\begin{abstract}
Telepresence robots offer presence, embodiment, and mobility to remote users, making them promising options for homebound K-12 students. It is difficult, however, for robot operators to know how well they are being heard in remote and noisy classroom environments. One solution is to estimate the operator's speech intelligibility to their listeners in order to provide feedback about it to the operator. This work contributes the first evaluation of a speech intelligibility feedback system for homebound K-12 students attending class remotely. In our four long-term, in-the-wild deployments we found that students speak at different volumes instead of adjusting the robot's volume, and that detailed audio calibration and network latency feedback are needed. We also contribute the first findings about the types and frequencies of multimodal comprehension cues given to homebound students by listeners in the classroom. By annotating and categorizing over 700 cues, we found that the most common cue modalities were conversation turn timing and verbal content. Conversation turn timing cues occurred more frequently overall, whereas verbal content cues contained more information and might be the most frequent modality for \emph{negative} cues. Our work provides recommendations for telepresence systems that could intervene to ensure that remote users are being heard. 
\end{abstract}






\maketitle

\section{Introduction}

\begin{figure}%[b]
  \centering
  \includegraphics[width=\columnwidth]{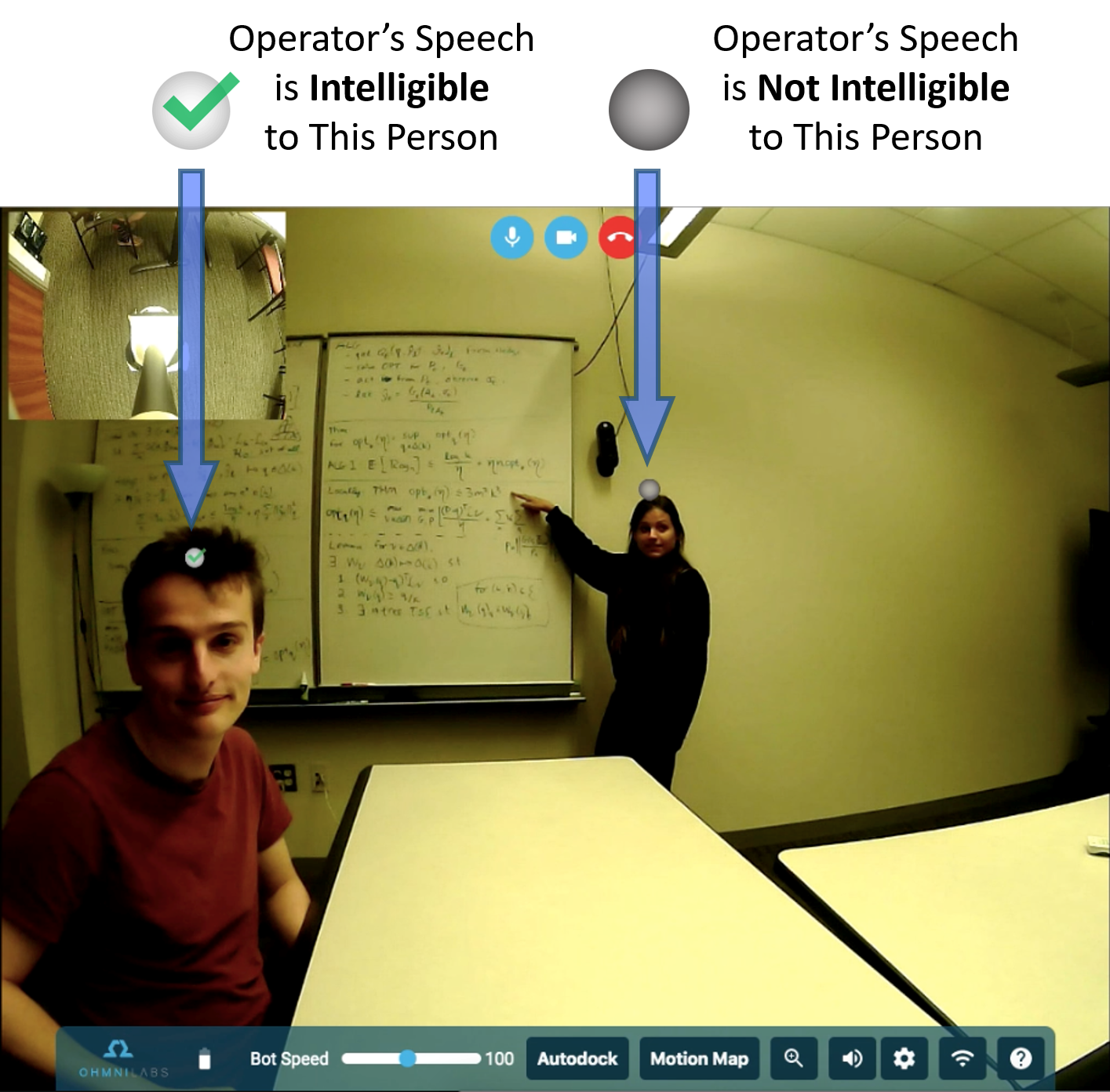}
  \caption{Screenshot of the Ohmni telepresence robot operator interface with enlarged views of the elements added by the speech intelligibility feedback system. The model has estimated that the closer person (seated at left) can understand what the operator has said, but the farther person (standing at center) cannot. }
  \Description{}
  \label{fig:volume-feedback-system}
\end{figure}

Telepresence robots are a promising option for K-12 students who are homebound and must attend class remotely~\cite{weiss_pebbles_2001,fels_telepresence_2001,newhart_virtual_2016}. A telepresence robot, also called a mobile remote presence (MRP) system~\cite{lee__2011}, allows the user to navigate and perform two-way videoconferencing in a remote environment~\cite{kristoffersson_review_2013}. Research has explored the use of telepresence robots in a wide variety of contexts including homes~\cite{beer_mobile_2011,tsui_exploring_2011}, hospitals~\cite{vespa_intensive_2007}, offices~\cite{lee__2011,venolia_embodied_2010}, museums~\cite{museum_robots_2014}, and professional conferences~\cite{neustaedter_beam_2016}. In 2001, in-school work showed the promise of telepresence robots for helping remote students to keep up with course material and contribute to classroom discourse~\cite{weiss_pebbles_2001,fels_telepresence_2001}. Subsequent research demonstrated that telepresence robots can minimize the effects of physical separation during school absences of K-12 students~\cite{newhart_virtual_2016}. Multiple studies have also shown ways for telepresence robots to provide better experiences than other distance learning technologies, albeit in university classrooms~\cite{bell_2d_2016,gleason_hybrid_2017,fitter_are_2020}.

Hearing each other clearly is a problem for users of telepresence systems, especially in the K-12 classroom setting. In a field study of a telepresence robot in a K-12 classroom, designers operating the robot mentioned audio and video communication quality as a key challenge~\cite{cha_designing_2017}: it took a great deal of effort to hear conversations over distance or through ambient noise, and the robot's volume setting was often at the wrong level. Speech intelligibility in classrooms is subject to classroom acoustics~\cite{bradley_intelligibility_2008} where voices can interfere with one another, making the target voice harder to hear~\cite{bronkhorst_cocktail_2000}. The student's home environment also impacts the intelligibility of their speech; not all students have access to a quiet space with a good Internet connection. The student's speech intelligibility also depends on the quality of their microphone and the robot's speaker. Telepresence robots have had audio-related issues in other applications as well; in the workplace, for example, operators can inadvertently speak too loudly~\cite{paepcke2011yelling,lee__2011}. 

Specifically, it is difficult for the telepresence robot operator to tell how loudly they should speak. Their voice is competing with other sounds in the classroom that are conveyed to them indirectly through the audio connection, and may even be filtered out as background noise. Additionally, viewing the distorted feed from a wide-angle robot camera on a relatively small screen makes it challenging for the robot operator to estimate the size of the classroom and the distance to a potential listener. As a result, the telepresence robot operator's speech might not be appropriately loud and well-enunciated for the classroom context, and they must rely on their listeners for signals about whether they are speaking at the appropriate volume to be heard and understood. 

Telepresence systems could help the robot operator to speak at an appropriate volume via several types of interventions. Sidetone, a technique used in the past in telephones~\cite{lane_regulation_1970}, appears not to work as well with telepresence robots~\cite{paepcke2011yelling}. Dynamic volume control techniques~\cite{tsui_exploring_2011} such as audio compression and ambient noise compensation~\cite{paepcke2011yelling} would require adjusting to the audio characteristics of the operator's speech and all the background sounds that could occur in their home. {\it We evaluated a different approach: giving the robot operator visual feedback about their speech intelligibility}. Our system
\makeatletter
\if@ACM@anonymous
    ~[citation omitted for anonymous review]
\else
    ~\cite{rueben2020increasing}
\fi
\makeatother
displays visual feedback on the operator's interface about multiple potential listeners based on a data-driven model of speech intelligibility. This system was deployed with two homebound K-12 students for several weeks---the first deployment of a speech intelligibility feedback system in the K-12 context. We present findings from subjective interview data and objective behavioral measures. 
% In the past, telephones played back the speaker's voice into their ear with a reduced volume to simulate hearing oneself speak; this technique is called \emph{sidetone}. 

%\begin{anonsuppress}
%\end{anonsuppress}

To provide the visual feedback described above, the telepresence system needs an estimate of the robot operator's speech intelligibility to a particular listener. A listener comprehension model providing such an estimate should include comprehension cues from the listener. People monitor their own comprehension levels from an early age~\cite{flavell_development_1981} and use various types of verbal~\cite{hirasawa2000new,meena2015automatic}, prosodic~\cite{hirasawa2000new,hirschberg2004prosodic}, and nonverbal~\cite{nakano2003towards,sahindal2020detecting} cues (such as gaze and facial expressions) to signal incomprehension. {\it We performed the first analysis of the types and frequencies of comprehension cues given by listeners to telepresence robot operators in K-12 classrooms.} We collected data from four in-the-wild deployments, including the two mentioned above.  %Our annotation and analysis of audio-visual recordings has implications for future telepresence systems (and perhaps social robots) that seek to automatically estimate listener comprehension of natural language utterances.

% should probably include environmental factors such as ambient noise levels and distance to the robot as predictors~\cite{takahashi2015case,hayamizu2014volume}, but cues of comprehension (or noncomprehension) from the listener would probably be important as well, for correcting the model's estimates. 

This work makes two complementary contributions---an evaluation of a speech intelligibility feedback system and an analysis of listeners' comprehension cues---as part of a larger study of how telepresence robots can help homebound K-12 students stay connected to their classrooms 
\makeatletter
\if@ACM@anonymous
    ~[six citations omitted for anonymous review]\else
    ~\cite{cha_my_2017,cha_designing_2017,fitter_evaluating_2018,fitter_comparing_nodate,rueben2020increasing,fitter_are_2020}\fi
\makeatother
. Both contributions are firsts for telepresence in the K-12 classroom, and our findings also provide insights for the other telepresence applications mentioned above: homes, hospitals, workplaces, museums, and conferences.

\section{Related Work}

\subsection{Feed-Forward Models of Listener Comprehension}
Dialogue systems need to estimate multiple variables about a user during a conversation, including their goal~\cite{chen_survey_2017}, the likelihood that they are about to end their turn~\cite{skantze2020turn}, and their emotional state~\cite{ma_survey_2020}. Models producing those estimates often use multimodal data~\cite{cambria_sentic_2013}. 

A speech intelligibility feedback system requires a model for estimating listener comprehension. Several such models have been developed. In the context of human-robot interaction, \citet{mead2014probabilistic,mead_perceptual_2016} developed a model of both listener comprehension and typical response volume based on user distance from the robot. \citet{hayamizu2014volume} and \citet{takahashi2015case} developed additional models of listener comprehension that included ambient noise in addition to distance. Comprehension might also depend on the listener's head orientation and prosodic features of the speaker's voice~\cite{davat_integrating_2018,davat_can_2020}.

This work presents an evaluation of a simple feed-forward listener comprehension model we developed specifically for K-12 students that uses both distance and ambient noise 
\makeatletter
\if@ACM@anonymous
    [citation omitted for anonymous review]\else
    \citep{rueben2020increasing}\fi
\makeatother
. The model was trained using data collected from the same telepresence robot system we used in this work, capturing its microphone, speaker, and audio transmission characteristics. The model does not estimate to whom the robot operator is speaking; instead, listener comprehension is calculated for each potential listener in the classroom.

\subsection{Speech Intelligibility Interventions for Telepresence Systems}
There are several ways to intervene when a listener comprehension model determines that a listener is unlikely to have understood the telepresence user's utterance. 

One approach is to automatically control the robot's output volume.  \citet{hayamizu2014volume} and \citet{takahashi2015case} trained two listener comprehension models to control a telepresence robot's output volume based on distance to the listener: one for ``normal conversation'' and another for ``secret talk'', so others besides the listener could not hear. \citet{paepcke2011yelling} used audio compression to soften loud utterances, but this approach is parameterized by the expected duration of the sounds needing attenuation so its effectiveness would depend on the noise characteristics in each operator's home. \citet{escotta_controle_2021} used fuzzy logic to increase the system's volume in response to background noise.

A telepresence system can also provide enhanced audio immersion to the operator. \citet{liu_auditory_2019} provided binaural auditory scene reproduction to the operator using the operator's head orientation and the 3D locations of sound sources.  \citet{paepcke2011yelling} evaluated sidetone---playing the robot operator's voice back to them with a slight delay---on telepresence robots and found that it did cause operators to talk more quietly, but the effect was smaller than in studies of audio-only interactions. They hypothesized that the visual component of the telepresence system interferes with the desired effect of sidetone. 

A third approach is to provide visual feedback to the operator. \citet{paepcke2011yelling} evaluated using graphical sound pressure level (SPL) feedback in the user interface, but robot operators tended to overlook this feedback. \citet{kimura_visual_2007} calibrated a similar feedback system to the distance to the listener, and additionally projected the same feedback onto the floor where the listener could see it.

This work presents an evaluation of an intervention described by 
\makeatletter
\if@ACM@anonymous
    [author(s) and citation omitted for anonymous review]\else
    \citet{rueben2020increasing}\fi
\makeatother
: an icon positioned above the face of each potential listener that indicates whether the robot operator's speech is loud enough to be intelligible to them. In contrast to the visual feedback used by \citet{paepcke2011yelling} and \citet{kimura_visual_2007}, our system displays feedback for each potential listener, and does so above each of their faces in the video image for visual salience. Unlike the prototype by \citet{kimura_visual_2007} however, our system gives no feedback to the listener.  

\subsection{Listener Comprehension Cues}
Models for estimating listener comprehension could be improved by detecting cues from the listeners about their comprehension levels and using that information to update the model's predictions. 

Research on establishing mutual understanding in human-robot interaction~\cite{kiesler2005fostering}---i.e., establishing \emph{common ground} via \emph{grounding}---has primarily focused on enabling the robot to understand the human~\cite{ros2010one,shridhar2018interactive,roesler2019evaluation}. In conversations via a telepresence robot, however, the opposite is also important: the robot should be gathering evidence about whether its operator's utterances are being understood by the remote listeners who are physically co-present with the robot. \citet{roque2009improving} developed a grounding model for a virtual agent equipped with a spoken dialogue system; for evidence of grounding their model used cues that included repeating an utterance, using the provided information, and failing to respond.

Other types of listener cues have been used to detect speech recognition failures (i.e., incorrect speech transcriptions) in spoken dialogue systems. Some failure detection models use the content of the user's response~\cite{hirasawa2000new,meena2015automatic}, including repetition of the same utterance, disconfirmations using ``no'' or ``not'', and making corrections by repeating an utterance with modifications~\cite{meena2015automatic}. Other work uses prosodic features such as tempo and duration of utterances and pauses~\cite{hirasawa2000new,hirschberg2004prosodic}. Face-to-face interactions with embodied virtual agents and physical robots offer additional cues such as gaze, gestures (e.g., head nods), facial expressions, and body motion~\cite{nakano2003towards,sahindal2020detecting}. Some systems have combined multiple modalities of information---e.g., dialogue and gestures~\cite{lucignano2013dialogue} or prosody and sentiment analysis~\cite{li2017emotion}. 

Our study is the first to catalog the types of multimodal comprehension cues given by listeners to telepresent students in the K-12 classroom context. We estimate which cue modalities are most common in this context to inform the design of telepresence systems that can provide timely interventions when the remote student is not being heard. 

\section{Model-Driven Feedback About Speech Intelligibility}
%speech intelligibility feedback to the third and fourth participants in this study. 

The development of our speech intelligibility feedback system has been described in detail by 
\makeatletter
\if@ACM@anonymous
    [author name(s) and citation omitted for anonymous review]\else
    \citet{rueben2020increasing}\fi
\makeatother
. The system provides the robot operator with real-time visual feedback designed to minimize distraction from the ongoing conversation. Determining to whom the robot operator is attempting to speak is avoided by providing feedback for every potential listener in the camera's field of view. As shown in Figure~\ref{fig:volume-feedback-system}, feedback is provided in the form of a small icon positioned over the top of each detected face in the camera frame. 

The model operationalizes speech intelligibility as speaking volume appropriateness to determine which of the two icons to display: intelligible (loud enough) or not intelligible (not loud enough). The model was trained on speech intelligibility data from three participants in a laboratory setting. Two variables were controlled and systematically varied: listener-robot distance and ambient noise volume. The following planar model %shown in Equation~\ref{eqn:volume-model} 
was found to be a good fit for the training data:
\begin{equation}
\begin{aligned}
    MinimumIntelligibleVolume = 0.01168*AmbientNoise + \\
    6.90635*ln(Distance) + 49.40575
\end{aligned}
\label{eqn:volume-model}
\end{equation}

The model outputs the minimum intelligible speaking volume (in decibels) for a given ambient noise level (in the native units of the robot's microphone) and listener-robot distance (in meters). We used an additional model trained on data collected in the lab to estimate the listener-robot distance from the listener's face height in pixels.

\section{Long-Term Deployments With Homebound K-12 Students}
This work is part of a larger project studying how telepresence robots can reduce the impact of extended school absences of homebound K-12 students. We deployed a telepresence robot system with four homebound K-12 students, referred to as P1, P2, P3, and P4, who used the robot to attend classes for several weeks while we collected a multimodal dataset. In this section we present the data collection process and then describe our multimodal dataset.

\subsection{Data Collection}
To obtain meaningful data beyond novelty effects in both homes and classrooms, we collected data for at least two weeks for each participant (see Table~\ref{tbl:participants} for exact durations) to assess how student operators received feedback about speech intelligibility and how students and teachers in the operators' classrooms provided it. 

Recruitment was extremely challenging because it required finding participants who were homebound from recruitment through to the end of the deployment. Nevertheless, it was important to capture data that reflected the diverse reasons students become homebound, along with their associated accessibility challenges and other individual differences. %Furthermore, the qualifying paricipant pool was exceedingly diverse, introducing novel challenges for both the telepresence technology and the data collection. 

The telepresence platform was the Ohmni robot by OhmniLabs; we enhanced the web-based interface with several modifications, including our speech intelligibility feedback system. Participants accessed the interface via Macbook Air laptops we deployed to their homes. The interface included a slider for adjusting the volume the robot used to play the participant's speech. Participants either used earbud headphones with built in microphones or the laptop's built-in microphone and speakers. 

%\subsubsection{Deployment Procedure}
Participants and their families were trained in their homes on how to use the telepresence robot, including the speech intelligibility feedback system. P1 and P3 needed family members to set up and log in for them, while P2 and P4 could do so independently. After the training, the robot was removed from the participant's home and placed in their classroom. The homebound students then attended one or more classes using the robot, subject to their class schedules and health constraints.

% The first login was often done with a member of the study team or an education technology specialist present to help troubleshoot any technical issues. The homebound student was then introduced to the class by the teacher. We also encouraged the teacher to choose someone in the classroom to help the participant with technical issues or to carry or push the robot. Sometimes this was the teacher themself, or a teacher's aide, or another student---the ``bot buddy''. 

Various technical issues arose, especially at the start of each deployment. P1, P2, and P4 experienced times when they could hear the classroom but could not be heard; for P1 this lasted the first few days of the deployment. Some audio-visual recordings were not started or ended correctly, causing data to be lost. For example, one of the two recordings was missing for 2 hours and 15 minutes of the 14 hours of recordings that were annotated.  %The study team acted as live technical support for the deployments, often fixing these problems remotely. 

%P1 had several days where the class couldn't hear him!
%P2: teacher can't hear
%P3: was muted briefly, maybe on the robot side
%P4: classmates can't hear, use a sort of sign language since P4 isn't showing their face

Most participants attended entire class sessions via the robot, except for P3, who usually attended for at most an hour, for health reasons. P2 and P4 attended multiple classes via the robot (see Table~\ref{tbl:participants}). Most participants also had some telepresence interactions outside the classroom, e.g., en route to the next classroom. 

%Surveys were administered to the participant and their teacher(s) and classmates. Some surveys were administered weekly, whereas others occurred just once or twice in a deployment, such as at the very beginning and the very end. 

At the end of the deployment period, the robot was retrieved from the school and the laptop from the participant's home. Participants, their families, and the teachers were compensated for participating in the study. Classmates were given a small gift for their participation. The project was approved by the Institutional Review Board (IRB), study number UP-18-00129.

Table~\ref{tbl:participants} provides the the details of the four participants' deployments. The speech intelligibility feedback system was developed during the P1 and P2 deployments and subsequently deployed with P3 and P4. Comprehension cues were annotated and analyzed for all four participants. P1 and P3 were difficult to understand: P1 was on a ventilator and spoke in brief bursts, and P3 had significant difficulty enunciating due to their medical condition. 

% Three of our survey questions contribute to the evaluation of the volume feedback system: 
% \begin{itemize}
%     \item ``I found the volume appropriateness feature distracting.''
%     \item ``I paid attention to the volume appropriateness feature.''
%     \item ``If I used a robot like this all the time in class, I would want to use the volume appropriateness feature.''
% \end{itemize}
% Only P3 was administered these questions; we mistakenly failed to administer them to P4.
\subsection{Multimodal Dataset}
We collected multiple types and modalities of data. 

Semi-structured interviews were conducted with participants and their caregivers at least once per week. P3 and P4, to whom we deployed the speech intelligibility feedback system, were asked about it at every interview. At the final interview, they were also asked for ways to improve the feedback system, e.g., by making it clearer or easier to use. 

Behavioral data were collected whenever the participant was logged into the robot via audio and video recordings from two perspectives. The camera and microphone on the homebound participant's laptop recorded the homebound participant's face and voice as presented to the classroom by the robot. The robot's camera and microphone recorded the classroom as it was presented to the homebound participant on their laptop screen. It was often difficult to understand what people in the classroom were saying, due to ambient noise or distance from the robot.
%Accordingly, both sides of every conversation were recorded. 

Sixty-seven hours of audio-visual recordings were collected: 14 hours from P1, 10 hours from P2, 15 hours from P3, and 28 hours from P4. 
%  (see Figure~\ref{fig:what-got-annotated})

\begin{table*}[]
\caption{Details of the four deployments.}
\begin{tabular}{p{3.7cm}|>{\raggedright}p{2.5cm}|>{\raggedright}p{2.5cm}|>{\raggedright}p{2.5cm}|p{3cm}}
\textbf{Participant}                   & \textbf{P1}              & \textbf{P2}         & \textbf{P3}              & \textbf{P4}                            \\ \hline
\textbf{Duration}                      & 5 weeks                  & 2 weeks             & 8 weeks                  & 5 weeks                                \\
\textbf{Calendar Days}                 & Mar 26--Apr 17, 2019     & Jun 3--Jun 13, 2019 & Oct 29--Dec 17, 2019     & Feb 21--Mar 13, 2020                   \\
\textbf{Grade}                    & High School                  & High School    & High School        & Middle School \\
\textbf{Classrooms}                    & Spanish                  & History, English    & Intensive Studies        & Math, Science, English, History, Lunch \\
\textbf{Feedback System?}                 & No                  & No              & Yes                  & Yes                                \\
\textbf{Days Attended}                 & 10 days                  & 7 days              & 21 days                  & 13 days                                \\
\textbf{Recorded Time on Robot}           & 14 hours                 & 10 hours            & 15 hours                 & 28 hours                               \\
\textbf{Difficulties Using the Robot?} & Yes: physical and speech & None                & Yes: physical and speech & Minor physical                        
\end{tabular}
\label{tbl:participants}
\end{table*}

\section{Data Analysis}
\label{sec:annotation}

The audio and video recordings described in the previous section were manually annotated for comprehension cues sent to the participant from their classmates and teachers. We also analyzed the recordings for speech intelligibility feedback system use and its effect on participants' use of the robot's volume slider.

% \subsection{Feedback System Usage and Volume Adjustments}
% The feedback system for speech intelligibility was supposed to be turned on by the participant via a button. It was not always turned on right at the beginning of the login session, however, especially because the participant needed to remember to turn it on again any time the webpage hosting the interface was refreshed. We manually watched the video recordings and recorded each time that the feedback system was turned on because this information was not logged.

% We also wanted to record volume slider adjustments to test whether the volume feedback system encouraged volume adjustments. A custom script was written to use template matching to identify whether the volume slider was displayed and, if so, the current volume level. This script was run on the recordings for P3 and P4 after the deployment. The script assumed that the Ohmni interface webpage was always displayed at full screen size with no other windows covering it; it is possible that some volume slider adjustments were missed if these assumptions were violated.

\subsection{Annotation Personnel and Process}
\label{ssec:annotation-process}
The annotation team annotated a subset of the audio-visual recordings because of the time-intensive nature of our annotation process: 14 hours and 20 minutes were annotated, or about 21\% of the total duration. Of these, 2 hours and 20 minutes (about 16\%) were co-annotated---i.e., annotated by both annotators to evaluate inter-rater reliability. Annotation assignments were distributed approximately evenly across the four participants and then across the weeks within each participant's deployment. Co-annotations were also distributed approximately evenly, except that none were assigned for P1, who spoke very seldom and therefore received very few comprehension cues. 
% As shown in Figure~\ref{fig:what-got-annotated}, a

% \begin{figure}
%   \centering
%   \includegraphics[width=.9\columnwidth]{what-got-annotated.png}
%   \caption{(placeholder caption)}
%   \label{fig:what-got-annotated}
% \end{figure}

We used the free annotation software ELAN. The two audio-visual recordings for each class session---of the participant from their laptop and of the classroom from the robot---were synchronized and displayed simultaneously in ELAN for the annotators. Additionally, a volume analysis of the participant's audio and an optical flow analysis of the robot's video feed were used to eliminate sections of the recording that did not contain participant speech or robot movement, respectively. 
% The different phenomena and variables to be annotated were displayed in ELAN as rows with drop-down menus with the options for each variable. 
% Annotators made multiple passes through each recording, identifying instances of the different phenomena on an initial pass and then returning to fill out the details. 

The recordings were annotated for participant speech, robot movement, comprehension cues, and classroom context. %Each protocol included an operational definition of that phenomenon (e.g., what counts as a comprehension cue) and instructions for evaluating the associated variables (e.g., determining whether a cue is positive or negative---see below). The protocols were updated throughout the annotation process in response to our decisions about cases in the recordings that were difficult to classify.
%with examples, especially borderline examples collected during the annotation process. 
%Each time an annotator identified one of these phenomena (e.g., an utterance spoken by the participant), they evaluated several variables for that instance (e.g., duration, topic, who started the conversation)---the protocol also contained the options for each variable, rules for deciding which to choose, and examples for each option.  
The annotation team consisted of two annotators, two annotation supervisors and protocol developers, and a team leader. %One of the annotation supervisors developed the protocol for annotating for comprehension cues and classroom context, and then supervised the annotation of those variables. The other supervisor was in charge of speech and robot movement annotations. 
%The team leader oversaw the development of the protocols and the annotation process.
For training, the annotators participated in protocol development and annotated test recordings. The two annotators' work was compared and then reviewed by both supervisors. 
%Corrections were discussed at team meetings so both annotators could learn from each mistake. 
% The annotators' insights contributed significantly to protocol development throughout the annotation process. 

%After completing their training, the annotators began annotating the chosen selections from the main dataset. 
%Their first set of assignments were co-annotations so we could continue comparing their work and discussing the same sections of the recordings at team meetings. 
% (see Figure~\ref{fig:what-got-annotated}) 

The annotation supervisors continued to review all annotations after the annotators completed their training and began annotating their assigned recordings. The annotators then re-watched any sections of video flagged by a supervisor and revised the annotation file in response to the comments. The responses were reviewed to ensure that all comments were properly addressed.

\subsection{Annotation of Comprehension Cues}
\label{ssec:cue-annotation}
\emph{A comprehension cue} was defined as any action or inaction by a potential listener (the cuer) in the classroom from which a reasonable robot operator would gather information about whether (or to what extent) that person heard them. We intended for the annotators to notice more cues than the participants themselves did, in order to catalog as many cues as possible. In addition to a text description of each cue, the annotators recorded: the cuer (teacher, TA, or classmate); whether the cue was verbal or nonverbal; whether the cue was positive (i.e., indicating that the cuer understood the participant), negative (i.e., indicating that they did not), or mixed (e.g., partial understanding); and an initial cue category that was later replaced by a final cue category, subcategory, and modality. %The initial cue categories were as follows: imperative cues were commands, e.g., to speak up or increase the volume; informative cues explicitly informed the participant that, e.g., the cuer did not understand what they said; affirmative cues were responses or actions that show understanding through affirming or complying with what the participant said; implicit cues include not responding, repeating the participant's utterance with mistakes, and leaning forward to hear better.  

\subsection{Development of Cue Categories and Labeling by Modality}
\label{ssec:categorization}
After the annotation process was complete, cue categories and subcategories were developed to fit the dataset of cues. The cue categories were developed iteratively between two of the authors: categories were added, merged, or split until every cue fit in at least one category and categories had minimal conceptual overlap. Each cue was assigned to only one category; cues that fit into multiple categories were assigned to the category that contained the most information about listener comprehension. %Each category was given a definition and every cue was assigned to a category. %the cues were organized into categories and then subcategories that were developed to fit the dataset better than the initial cue categories did.
%by the cue annotation supervisor and the annotation team leader 
Next, all the cues assigned to each category were reviewed again together, and some but not all of the categories were split into subcategories to better fit these cues. The subcategorization process resulted in the merging, splitting, deleting, and adding of categories as well as subcategories. 
% Some of these subcategories were inspired by the verbal/nonverbal or positive/negative labels from the initial cue annotation---e.g., the ``No Response'' category has a verbal subcategory for cuer utterances that ignore the participant, and a nonverbal subcategory for cuer silence. 

%\subsection{Labeling Subcategories by Modality}

We specified which modality of data would be needed to detect and interpret the cues in each subcategory. Focusing on modalities kept the analysis agnostic to different choices of algorithm or software implementation. The first three top-level modalities we chose were audio, video, and robot movement and settings. Audio was subdivided into verbal and nonverbal voice data (no non-vocal audio cues were found in this study). Next, the verbal data submodality was further subdivided into cues based on key words or phrases and cues based on one or more sentences. Video was subdivided into four submodalities: (1) head, eye, and face (e.g., head movements, eye gaze, and facial expressions); (2) hands; (3) other body pose (requiring the trunk or legs, such as leaning towards the robot); and (4) activity (i.e., higher-level actions such as walking or picking something up). 
%(see bottom part of Table~\ref{tbl:categories-and-modalities}) 
%Each of these comes from a different sensor or set of sensors---namely, a microphone, a camera, and the robot's wheel encoders and log files. 

The fourth and last top-level modality was conversation turn timing. While turn-taking in spoken dialogue is often analyzed using audio or video data, we chose to consider it as a separate modality because it focuses on the timing of the conversation instead of its content.

\section{Evaluation of the Speech Intelligibility Feedback System}
%The speech intelligibility feedback system appears to only have been used once: by P4, and successfully. This section details what can be learned from this event, 
This section presents our findings about the speech intelligibility feedback system from data collected in deployments to P3 and P4.

\subsection{Interview Findings}
Both P3 and P4 reported that the speech intelligibility feedback system did not distract from participating in classroom activities, but did not use it very often for several reasons. One was misunderstanding: as late as Week 6, P3's parent (who helped P3 use the robot's interface) reported an incorrect belief that the feedback indicates who is speaking in the classroom. Also, both P3 and P4 reported a belief that the feedback system increased network latency. This sometimes led P3 and their parent to leave the feedback system turned off. 
%During Week 3, P3's parent said they (the parent) did not pay any attention to it. 
%The interviewer corrected them. 

Nonetheless, both participants felt the speech intelligibility feedback system was well-suited for certain applications. P3's parent said it could help to ensure that the teacher could hear P3 over other conversations in the classroom---a common problem for P3, who spoke a lot and often wanted the teacher's attention. P4 wanted to use it to have small, private conversations with friends without being overheard by teachers or other students. Late in Week 6, P4 successfully used the feedback system in this way during a conversation with a classmate they were befriending. P4 also commented that the system could be useful for giving class presentations.

\subsection{Behavioral Findings}
Analysis of the video recordings revealed that adjustments to the volume slider in the robot's web interface by participants or their family members were very rare. P3 and their family members adjusted the volume slider 8 times and viewed it 2 additional times in the 10 hours of recorded video that we analyzed (out of 15 hours total). P4 adjusted the volume slider 9 times, 6 of which were unrelated to any utterance from or to P4, in the 23-½ hours of recorded video that we analyzed (out of 28 hours total).

We did not find evidence that the speech intelligibility feedback system encouraged more volume slider adjustments. In fact, for both participants there were fewer adjustments or views per hour when the feedback system was on than when it was off. P3 had 6 adjustments or views in the 4-½ hrs of video we analyzed when the feedback system was off and 4 in the 5-½ hrs when it was on. P4 made 8 adjustments in the 17-½ hrs of video we analyzed when the feedback system was off and just 1 in the 6 hrs when it was on. %Gathering more weeks of data from these two participants may have revealed that there is in fact a positive effect, but given these data it would probably not be a very large one. 

\section{Analysis of Comprehension Cues}
As described in Section~\ref{ssec:annotation-process}, 14 hours and 20 minutes of audio-visual recordings from all four participants were annotated for comprehension cues given by listeners. In this section we present (1) the agreement levels between the two annotators and results from (2) the initial annotation, (3) categorization, and (4) grouping by modality of the 774 cues we identified.

\subsection{Inter-Annotator Agreement}
\label{ssec:inter-annotator-agreement}
Of the 774 total cues, 231 were in the sections of the recordings that were annotated by both annotators so we could compare their annotations. Only 74 (32\%) of these 231 cues were recorded by both annotators; 89 were only recorded by the first annotator and the remaining 68 were only recorded by the second annotator. This low agreement is not problematic if each annotator was attentive to different types of cues, thereby together finding more cues than separately. It might also suggest, however, that annotating for all types of cues at once was too difficult for the annotators. 

There is evidence that our annotators had lower agreement when it was difficult to tell what actions and inactions were related to a participant's utterance. This difficulty was made worse by any network latency causing delays between participant utterances and responses from the classroom, which also could have made it difficult for the annotators to assess conversation turn timing. 

Additionally, many of P3's utterances were very difficult to understand and spoken to nobody in particular or out of turn (e.g., when the teacher was speaking to the whole class). These factors often made it difficult to tell whether an action or inaction by a teacher or student was related to an utterance by P3. In the co-annotated sections of video for P3, 41 cues that we later placed in the ``Beginning to Speak at an Appropriate Time'' category (see Table~\ref{tbl:categories-and-modalities}) and 19 that we placed in the ``Silence When a Response was Warranted'' subcategory were recorded by the first annotator but not the second annotator. These 60 cues account for about two thirds of the cues that only the first annotator recorded in the co-annotated video sections across all participants. Inspection of these cues reveals that it is often difficult to tell whether the cuer's appropriately timed utterance or lack of response is related to the most recent utterance by P3. 

Subcategories that more obviously met our definition of a comprehension cue (see Section~\ref{ssec:cue-annotation}) had much higher levels of agreement, such as ``Repeating What the Participant Said'' (7/8 = 88\% agreement) and ``Answering the Participant's Question'' (7/9 = 78\% agreement). 
%This indeed seems to have contributed to the low agreement between our annotators: i
% It is therefore difficult to decide whether they should be counted as cues. 
%Cues (perhaps of certain types) may have been missed by our annotators, especially if the first factor mentioned above (i.e., differential attention to different types of cues) was in effect. Therefore, the cues we have identified are useful for creating a preliminary taxonomy of cue types (i.e., our categories and subcategories) that occur in response to utterances by a telepresent student in K-12 classrooms, but this taxonomy is not necessarily complete. Also, the ratios of cue counts between subcategories and modalities might not be very reliable; in this report we focus on large differences and discuss possible threats to validity.  
Agreement was also high for labeling cues as positive, negative, mixed, or of uncertain valence: 60 (81\%) of the 74 cues recorded by both annotators were given the same label. 

Utterance counts---used in the next analysis to calculate cue frequency as average cues per utterance---also had relatively high agreement: 318 (80\%) of the 398 utterances recorded by either annotator were recorded by both. Most of the disagreement was for P3, who often laughed or made other sounds that may or may not have warranted a response. 

\subsection{Cue Frequency; Positive and Negative Cues}

\begin{table}[b]
    \centering
    \includegraphics[width=.95\columnwidth]{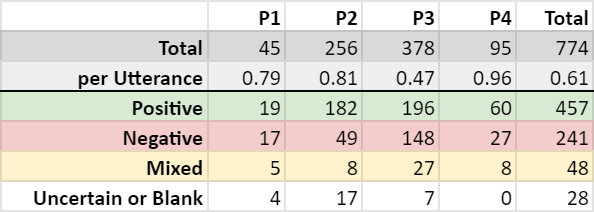}
    \caption{Summary of annotated comprehension cues.}
    \label{tbl:cue-totals}
\end{table}

Table~\ref{tbl:cue-totals} summarizes the comprehension cue annotation results before categorization. The large majority of the cues were for utterances by P2 or P3, who spoke much more often than P1 or P4 did. Compensating for this difference revealed that we did not notice more than 1 cue \emph{per utterance} on average for any of the four participants. 

%As discussed above in connection with inter-annotator agreement, P3 had an especially low rate of cues because they often made utterances with an unclear target and when the class was paying attention to the teacher; we often chose not to annotate a cue because it was unclear whether classmates did not respond to P3 because they did not hear their utterance, or because they were ignoring them to pay attention to the teacher. 

We noticed more positive cues (indicating comprehension) than negative cues (indicating a lack of comprehension) for all four participants. However, P1, who had audio issues for several days, and P3, who was difficult to understand and often spoke to the entire class out of turn, had higher proportions of negative cues than P2 or P4 did.

\subsection{Final Categories and Subcategories for Cues}
Table~\ref{tbl:categories-and-modalities} shows the categories and subcategories we developed to fit the cues. There were 15 categories and 31 subcategories. Seven of the categories had no subcategories, so there were 38 cue types (i.e., 7 categories without subcategories and 31 subcategories) that were labeled by modality. Some subcategories contained just a few cues, and two contained none---``Cuer Action Does Not Match Participant's Utterance'' and ``Reducing Classroom Noise to Hear Better''---but were included because they are possible.
%, as well as how these were categorized by cue modality (e.g., audio, video)

\begin{table}[]
    \centering
    \includegraphics[width=\columnwidth]{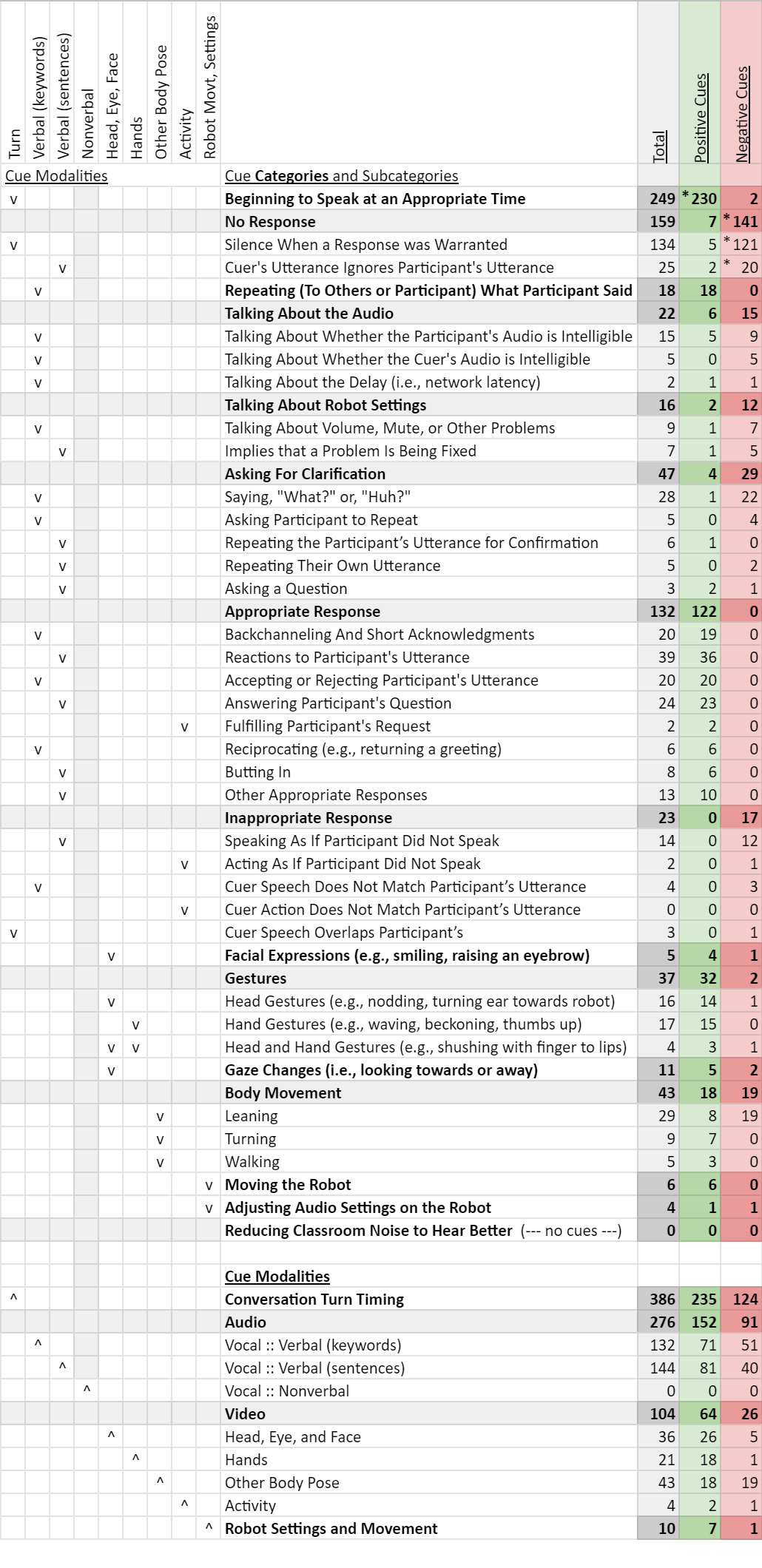}
    \caption{Comprehension cue counts by category and subcategory (top part of the table), and by modality (bottom part). Assignment of subcategories (and categories without subcategories) to modalities is shown on the left. Categories with subcategories and modalities with submodalities are colored gray, as is the column for the Vocal :: Nonverbal submodality because no subcategories were assigned to it. Numbers preceded by a star (*) were affected by P3's unique behavior, as discussed in Sections~\ref{ssec:inter-annotator-agreement} and~\ref{ssec:modalities}.}
    \label{tbl:categories-and-modalities}
\end{table}

%% CODE FOR SPLITTING A FIGURE ACROSS TWO PAGES
%\begin{figure}[!b]
%    \centering
%    \includegraphics[width=\columnwidth]{categories-and-modalities_1.PNG}
%    %\label{fig:arm1}
%    \caption{$Q^{*}$ values for different arms}
%\end{figure}%
%\begin{figure}[ht]\ContinuedFloat
%    \centering
%    \includegraphics[width=\columnwidth]{categories-and-modalities_2.PNG}
%    \label{fig:arm1}
%    \caption{$Q^{*}$ values for different arms}
%\end{figure}

\subsection{Cue Modalities}
\label{ssec:modalities}
For both positive and negative cues, the most frequent cue modality was conversation turn timing, followed by audio, and then video; the least frequent was robot movement and settings (see bottom of Table~\ref{tbl:categories-and-modalities}).

Conversation turn timing cues may have been the most frequent by an even wider margin if we had allowed each cue to be placed in more than one subcategory. The cues in every subcategory in the audio modality were likely to also belong in the ``Beginning to Speak at an Appropriate Time'' subcategory---i.e., to include information about listener comprehension in their timing as well as in their verbal content. As described in Section~\ref{ssec:categorization}, we categorized these cues according to their verbal content because it contained more information about listener comprehension than timing did. %We usually categorized cues by which part gave the most information about listener comprehension, so many cues that also displayed proper conversation turn timing were categorized according to their verbal content. 

If P3 is excluded, however, then the audio modality had the most \emph{negative} cues, which are important for indicating when the telepresence system needs to intervene. As noted earlier, P3 often spoke unintelligibly or out of turn, prompting their teacher and classmates to ignore these utterances. This inflated the number of negative conversation turn timing cues: 105 (85\%) of the 124 \emph{negative} conversation turn timing cues were categorized as ``Silence When a Response was Warranted'' or ``Cuer's Utterance Ignores Participant's Utterance'' in response to an utterance from P3. The audio modality had the most negative cues for each of the other three participants, totaling 48 (52\%) of the 93 negative cues they received.  %, , in applications where negative cues are of primary interest because they warrant an intervention by the telepresence system, the Audio :: Vocal :: Verbal cue modalities would be the most numerous for detection. 
% 114 (30\%) of the 386 conversation turn cues---

\section{Discussion}

\subsection{Interventions for Telepresence Operators}
Our participants may have simply spoken more loudly or softly as the need arose, removing the need to frequently adjust the robot's volume slider. The characteristics of the audio system may have still needed adjustment, however, for participants to achieve the desired range of volumes without difficulty. Personalization might also be needed---e.g., students with difficulties speaking loudly (like P1) or clearly (like P1 and P3) could have their voices amplified. The best calibration for a particular student might include nonlinear volume scaling, equalization, and compression. 
%Perhaps a good calibration would be for an utterance by the robot operator at a certain volume to be rendered by the robot at the same volume (or slightly higher to account for poor quality audio transmission) as if the operator had been speaking at that volume in class. 

Future work may address the development of a system for such calibration, to be used less frequently by adjusting more parameters than a single, master volume slider does. Related research would study how telepresence operators adjust their \emph{speaking} volume while interacting in the classroom and using other interventions such as sidetone.

Both P3 and P4 spent time talking with teachers or classmates about the network latency while using the robot. The latency level could instead be displayed to the operator and their teacher or classmate as part of another feedback system. For example, the system could display when a speaker's utterance finishes playing on the other end of the remote connection. %The feasibility and usefulness of such a system need to be evaluated. 

\subsection{Detecting Comprehension Cues}
We developed 38 subcategories (and categories without subcategories) to describe the diversity of cues in our K-12 classroom setting, but 54\% of the cues were in just three of them: ``Beginning to Speak at an Appropriate Time'' (32\%), ``Silence When a Response was Warranted'' (17\%), and ``Reactions to Participant's Utterance'' (5\%). In this section we discuss how a telepresence system could detect a large share of the available comprehension cues with a relatively simple detection system.

\subsubsection{Choosing the Modalities to Detect}
Cues in different modalities can carry different amounts of information. The subcategories in the audio modality, for example, often communicate \emph{how much} the cuer understood, whereas cues based on conversation turn timing only communicate whether the cuer could tell when the participant's turn ended. 

A single natural language understanding (NLU) system could detect cues in both the audio and conversation turn timing modalities. Even an NLU system that can only perform keyword detection and end-of-turn prediction could detect a large portion of the negative cues identified in our study (see Table~\ref{tbl:categories-and-modalities}). 

%The video and robot settings and movement modalities were rarer for our participants, which suggests they might not be worth pursuing except in application contexts wherein cues in those modalities are expected to be much more frequent. For example, a conversation via sign language would probably yield many more video cues, and a robot settings cues would probably be much more common when someone is trying to fix the robot. 

\subsubsection{Detecting Cues in the Most Common Modalities}

Our work identifies the most common modalities of comprehension cues used during our telepresence deployments in K-12 classrooms: conversation turn timing cues, keyword-level verbal cues, and sentence-level verbal cues. 

Cues in these modalities could be detected by real-time telepresence systems.  Specifically, designers could draw upon work towards fluent turn-taking in spoken human-robot interactions as reviewed by Skantze~\cite{skantze2020turn}. Other work used timing information such as tempo and duration of responses and silences to directly predict misunderstandings~\cite{hirschberg2004prosodic,hirasawa2000new}. Our work revealed that it can be difficult to determine whether a potential cue is in response to a particular utterance, however, especially when the the utterance is out of turn or difficult to understand (see Section~\ref{ssec:inter-annotator-agreement}). 

Models have also been trained using key words and phrases to detect features such as repetition of the same thing, disconfirmations via ``no'' or ``not'', and making corrections by repeating something with certain words changed~\cite{meena2015automatic}. Finally, spoken dialogue systems can understand utterances at the sentence level~\cite{devault2014simsensei} and there has been recent work on understanding more difficult utterances such as indirect speech acts (e.g., ``could you get me a coffee?'')~\cite{briggs2017enabling}.

\subsection{Improving the Cue Annotation Process}
Inter-annotator agreement for comprehension cues can be improved in multiple ways. Our cue annotation process was very difficult for our annotators. A better approach would involve multiple passes through the recordings, focusing on one cue modality (e.g., hands, face, posture, verbal content) at a time, for one potential listener at a time. However, this would take even more time than our already exceedingly time-intensive annotation process did. Alternatively, researchers could focus on cues in just one modality (e.g., conversation turn timing) or on dyadic interactions with only one potential listener.  
%For future studies that are also interested in all possible types of cues in noisy, multi-speaker environments like the K-12 classroom, 

\subsection{Limitations}
The evaluation of our speech intelligibility feedback system was based on a single prototype system trained on data from three people in the lab; the two participants who experienced it reported that they rarely attended to the feedback. While the self-reports showed that our system could be useful for private conversations and class presentations, alternative designs merit exploration.

The low inter-annotator agreement suggests that some cues, or even entire new categories of cues, were missed by our annotators. The relative frequencies we reported for different types of cues may also have been impacted, especially for certain cue types during P3's deployment, as discussed in Section~\ref{ssec:inter-annotator-agreement}. 

Additionally, the relative frequencies of different types of cues likely depend on contextual factors like the number and physical locations of potential listeners, and whether it is acceptable for the operator to speak. Therefore, the relative frequencies of cues in different categories and modalities could differ significantly between K-12 classrooms and compared to non-classroom environments.

\section{Conclusions}
We presented the first findings about speech intelligibility feedback to homebound K-12 students using a telepresence robot to attend class remotely. We analyzed a multimodal dataset collected from four long-term, in-the-wild deployments. 

Our analyses revealed that the two participants who evaluated the speech intelligibility feedback system rarely attended to it; future work should study how students speak at different volumes instead of adjusting the robot's audio settings. Future feedback systems should visualize network latency and help students calibrate more than one volume parameter. 
% but they gave encouraging evaluations of the system and helped us identify several directions for future research. 

By annotating and analyzing over 700 multimodal comprehension cues, we developed 15 cue categories and 31 subcategories. For our four participants, conversation turn timing and verbal content were the most frequent cue modalities. Conversation turn timing cues occurred more frequently overall, whereas verbal content cues contained more information and might be the most frequent modality for negative cues. Future telepresence systems would benefit from detecting and interpreting cues in the one or two richest modalities and using them to ensure that remote users are being heard, either by automatically adjusting the audio settings or by providing other interventions via the user interface.

\begin{acks}
This work was supported by: an NSF NRI grant for ``Socially Aware, Expressive, and Personalized Mobile Remote Presence: Co-Robots as Gateways to Access to K-12 In-School Education'', NSF IIS-1528121; the New Mexico Space Grant Consortium (NMSGC), NASA grant \# 80NSSC20M0034; and the New Mexico State University Office of the Vice President of Research and Graduate School (NMSU VPRGS). OhmniLabs generously provided the Ohmni robot platform as well as excellent technical support. We also acknowledge Shashank Saurabh for helping us detect when the speech intelligibility feedback system was on, and Christopher Birmingham and Frank Bernieri for helping with the literature review. The deployments would not have been possible without Prof. Gisele Ragusa, a whole host of people with the Long Beach Unified School District, the teachers and TAs, the participants, and their families.   
\end{acks}

%\bibliographystyle{ACM-Reference-Format}
%\bibliography{main.bib}
%\input{output.bbl}

%\appendix

%\section{An Appendix Section}
\balance

\end{document}